# Investigating AI's Challenges in Reasoning and Explanation from a Historical Perspective

Benji Alwis
Bootham School
York
UK

## Introduction

Imagine a scenario where an autonomous vehicle has been trained on an extensive dataset containing diverse driving situations, including numerous instances of pedestrians using designated crosswalks. This Artificial Intelligence (AI) model is designed to excel at recognizing crosswalks and navigating around pedestrians, prioritizing their safety. It has learned from this data that pedestrians typically cross the road at these marked locations, and it can discern the visual cues associated with crosswalks while accurately identifying pedestrian movements within them.

However, when faced with a novel and unfamiliar crosswalk, the AI system may encounter difficulties in accurately assessing the pedestrians' intentions. For instance, if a traffic signal at the crosswalk malfunctions, leading to confusion among pedestrians, the AI model might struggle to grasp the complexities of the situation and respond appropriately. This lack of the AI model's ability to understand and apply the underlying causes and effects in unfamiliar or unexpected situations could result in the autonomous vehicle failing to yield to pedestrians at crosswalks in specific circumstances. Such lapses in judgment could potentially lead to perilous situations, endangering the safety of both pedestrians and vehicle passengers.

The primary issue at hand lies in the AI model's deficiency when it comes to causal reasoning, the ability to deduce outcomes based on cause-and-effect relationships. The neural network-based approach, which has powered significant recent advancements in AI, excels at identifying correlations within extensive datasets. However, it often falters in terms of interpretability and rational deduction, serving as a barrier to developing AI systems with human-like capabilities.

Hence, the imperative lies in bolstering AI with the capacity for causal reasoning, enabling it to address "what if?" and "why?" inquiries effectively. This becomes particularly crucial in safety-critical domains where understanding the underlying causes and potential consequences of actions is paramount. This essay aims to delve into the roots of this deficiency through a historical lens by tracing the evolution of AI. While this drawback has been scrutinized from a technological standpoint (Knight; Schölkopf), it is noteworthy that, to the best of my knowledge, this represents the first attempt to dissect it from a historical perspective.

Building upon the argument put forth by Pinch and Bijker in their seminal work "The Social Construction of Facts and Artifacts," which contends that "technological determinism is a myth that results when one looks backwards and believes that the path taken to the present was the only possible path" (Bijker 28), I hold the view that the evolution of artificial intelligence is a complex process shaped not only by the inherent properties of the technology but also by an interplay of societal factors and human agency. Over the years, the development of technology has been profoundly influenced by a tapestry of social, cultural, and political forces. The framework of the "social construction of technology" (SCOT) offers a valuable lens through which we can understand this complex relationship between technology and society, with Pinch and Bijker laying its foundational stones. As explored later in this essay, the interplay between technology and society has played a pivotal role in shaping the trajectory of a nascent field like Artificial Intelligence during the pivotal period from the 1950s to the 1970s. Anderson and Rosenfeld elucidate some of the driving forces behind these dynamics, asserting that "bringing a new field into existence involves the participants in a bitter and sometimes brutal Darwinian struggle for jobs, resources, and reputation" (Anderson viii).

In this essay, I will adopt SCOT as the primary theoretical framework for dissecting the historical evolution of Artificial Intelligence. Throughout this analysis, I will leverage two fundamental principles integral to the SCOT framework.

The first critical factor examined is interpretive flexibility (Bijker 40). It suggests that technological innovations can be understood in various ways by different individuals or groups, leading to diverse perspectives and potential outcomes. This concept is a valuable tool for analyzing technology-related debates influenced by different social groups and their interests. It allows parties to frame technology differently, emphasizing certain aspects while downplaying others to advance their arguments. Moreover, it can be used strategically to evoke emotional responses and influence opinions.

In technology-related arguments, exploiting interpretive flexibility is a strategic communication technique that shapes the narrative around technology and its implications, influencing perceptions, opinions, and decisions.

The second critical factor is the role of relevant social groups in technology development and success. Various stakeholders, including engineers, designers, users, and policymakers, play a crucial role in negotiating and influencing the features and functions of technology. SCOT emphasizes the relationships and dynamics among these social groups and how they interact with the broader social context to understand how technology evolves and integrates into society.

In this essay, I delve into the evolutionary journey of Artificial Intelligence during its formative phase, with a particular focus on its inception and the pivotal factors that have steered its developmental course. Merton's insightful observations shed light on the dynamics that govern the emergence of nascent scientific disciplines and the roles played by both insiders and outsiders in molding these domains (Merton 10). He notes that established scientists, often referred to as "insiders," typically enjoy a greater share of recognition and resources for their contributions in comparison to newcomers, aptly labeled as "outsiders." This disparity in recognition can give rise to a self-perpetuating cycle where established researchers continually accrue more opportunities, consequently fostering further progress and triumph in their field. The privileged status of insiders affords them the influential power to shape the trajectory of the discipline. They exercise this influence through their sway over critical determinants such as funding allocation, the establishment of publishing norms, and the acceptance of innovative concepts. Conversely, outsiders may encounter formidable obstacles when endeavoring to gain acknowledgment for their ideas, secure essential resources, and establish their credibility within the field. Nonetheless, they bring to the table a breath of fresh perspectives and innovative concepts, challenging the prevailing paradigms and making indispensable contributions to the evolutionary path of the new field.

Merton's perspective underscores the pivotal role insiders play in delineating the contours of an emerging field, not only through their individual research endeavors but also through their collaborative ventures and interactions with fellow researchers. Moreover, external factors such as societal demands, technological breakthroughs, and interdisciplinary cooperation also exert substantial influence in molding the boundaries and the trajectory.

Actor-network theory (ANT), a concept pioneered by Bruno Latour, provides a framework for comprehending the genesis and legitimization of emerging scientific domains. ANT focuses on the web of interactions and collaborations involving an array of diverse actors, ranging from scientists and technologies to instruments, funding agencies, institutions, and even non-human elements. Within this framework, the alignment and coordination of these diverse actors are central to the development of shared interests and objectives (Latour 369). Unlike conventional perspectives that view scientific knowledge as the outcome of objective discovery, ANT posits that knowledge creation is a collaborative and co-constructive process involving both human and non-human entities. In essence, it asserts that the emergence of a nascent scientific field is characterized by a dynamic interplay of negotiations, controversies, and alliances among these multifaceted actors. According to actor-network theory, the conventional boundaries between scientific disciplines are not rigidly defined but are continuously subject to negotiation and construction. This perspective underscores the fluid nature of scientific boundaries, which are constantly evolving as a consequence of the interactions and negotiations among the various actors within the network.

Seidenberg characterizes the history of Artificial Intelligence research as a saga reminiscent of a long-running soap opera, populated by a cast of characters that may appear somewhat peculiar, even by the standards of academia (Seidenberg 122). In his vivid portrayal, this narrative unfolds as an extraordinary tale, replete with all the elements of a gripping drama – encompassing moments of tragedy, hubris, irony, humor, acrimonious intellectual battles, and, in a few instances, even the figurative "corpses" of ideas or projects that did not stand the test of time.

Alexander further delves into the underlying causes of these frictions and tensions within Artificial Intelligence research (Aleksander 29). He underscores that, in contrast to the linear and celebrated narrative of Watson and Crick's DNA discovery, the field of Artificial Intelligence is marked by a profusion of diverse and sometimes competing techniques and analytical methods. This inherent complexity, he contends, gives rise to a host of pressures and conflicts within the community of researchers. Among the significant catalysts of discord, Alexander identifies the relentless pressures to deliver tangible products, secure essential funding, the allure of mathematical rigor and analysis, and the desire to claim credit for innovative breakthroughs. These factors collectively contribute to the multifaceted and, at times, tumultuous nature of Artificial Intelligence research, creating a fabric of narratives that echo the human dynamics and motivations at play within the scientific arena.

The history of Artificial Intelligence research, as described by Seidenberg and analyzed by Alexander, emerges as a captivating narrative mixed with all the hallmarks of a compelling storyline. It is a testament to the complexity of scientific exploration and the multitude of forces that shape the evolution of ideas in this ever-evolving field.

**The Cybernetic Age: A New Frontier of Progress**

The emergence of AI owes its inception to the efforts of a diverse array of exceptionally skilled and intellectually assured individuals. This inclusive group encompassed mathematicians, electrical engineers, psychologists, and neuroscientists. While certain members of this cohort never fully integrated into the mainstream of AI research, their contributions wielded remarkable influence in shaping the initial trajectory of the field.

Amidst the backdrop of the Second World War, the significance of collaborative problem-solving soared to new heights. The pioneering collective of researchers whose collaborative efforts paved the way for the first wave of AI was composed of notable mathematicians like Norbert Wiener and John von Neumann, engineers including Julian Bigelow and Claude Shannon, neurobiologists Rafael Lorente de Nó and Arturo Rosenblueth, neuropsychiatrist Warren McCulloch, along with the unconventional genius Walter Pitts, who lacked formal qualifications (Heims 11). Although often referred to as the cybernetics group, it was not until 1947 that this collective solidified its identity as a distinct scientific field. Cybernetics, a term coined by Norbert Wiener, delved into the exploration of control, communication, and regulation principles in both natural and artificial systems. During this period, the emphasis shifted towards the human sciences, prioritizing pragmatic problem-solving over abstract musings (Heims 1). Heims has portrayed Wiener as 'the dominant figure' within the cybernetics group discussions, highlighting his role as a brilliant visionary and provocateur of innovative ideas (Heims 206). Wiener's view was that intelligence materializes through the complex processing of information facilitated by feedback mechanisms. This position stood in contrast to some of the prevailing notions of his era, including Sigmund Freud's theory, which suggested that the mind primarily orchestrates biological energies (Crevier 28). Wiener's inclination to integrate psychology into the framework of cybernetic concepts diverged from the approach of another influential figure within this group, John von Neumann (Edwards 240). Despite the divergent approaches they adopted, the cyberneticians forged a potent cluster of influential thinkers during their time. The history of science shows that, particularly within human sciences, such elite groups played an instrumental role in shaping consensus on priorities, leveraging their collective resources and prestige to propel research agendas forward. The historical records of the Macy conferences on cybernetics, a sequence of multidisciplinary gatherings supported by the Macy Foundation and convened from 1946 to 1953, show the pivotal role undertaken by this collective during that era (Heims 12). This stands in stark contrast to the divisions and conflicts explored in subsequent sections of this essay.

Within the group of researchers who infused cybernetics into their initial theories of intelligence, there were those who embarked on a mission to replicate the complex mechanisms of the brain. Their strategy involved the emulation of individual neurons through the use of electrical components. Neuropsychiatrist Warren McCulloch, a prominent figure among the cyberneticians, had been contemplating hypothetical engineering components designed to emulate the workings of the human mind and brain. He was known for his interdisciplinary approach, drawing insights from biology, mathematics, and philosophy to understand the brain and intelligence. In 1942, he met Walter Pitts. Upon their encounter, McCulloch, known for his benevolent nature, warmly offered him accommodation in his own residence, recognizing Pitts' homelessness (Anderson 3). Their collaborative efforts culminated in the creation of the McCulloch-Pitts (M-P) model, a computational framework depicting artificial neurons. This seminal innovation not only underpinned the inception of neural network theory but also cast a transformative influence on the landscape of computational neuroscience. The model's overarching objective was to encapsulate the fundamental operations of biological neurons alongside their prowess in information manipulation. Central to the M-P model was the introduction of artificial neurons configured as binary threshold units. These units took binary inputs, applied weighted connections, and produced binary outputs based on a threshold function. While the model was a simplification of real neural behavior, it demonstrated the potential for mathematical modeling of neural processes and information processing. It laid the foundation for subsequent developments in neural network theory and paved the way for the exploration of learning algorithms and more sophisticated neural network architectures. According to Jerry Lettvin, often considered the third vertex in the triangle of this collaboration (Kelly 55), Walter Pitts was a mere 17 or 18 years of age when the renowned McCulloch-Pitts paper titled "A Logical Calculus of Ideas Immanent in Nervous Activity" made its debut in the Bulletin of Mathematical Biophysics in 1943 (McCulloch). Lettvin has subsequently stated that

> *In no uncertain sense, Pitts was the genius of the group. He was also personally a very unhappy person. He was absolutely incomparable with the amount of knowledge he had (Anderson 9). To him the world was connected in a very complex and wonderful fashion. At the same time he was very very opposed to having his name known publicly, so much so that when they offered him the doctorate at MIT if he would just sign his name or translate a page from the German which he did very easily, he refused. Later on when they offered him a faculty position if he would just sign his name to a document, he refused* (Anderson 9).

In 1951, a collaborative group comprising McCulloch, Pitts, Lettvin, and Pat Wall presented themselves as a unified front to MIT. Notably, during this timeframe, Norbert Wiener, who had by then gained widespread recognition as a pioneer in cybernetics, had risen to a position of considerable influence. Upon his affiliation with MIT, McCulloch willingly relinquished his full professorship, accepting instead the post of research associate along with a modest apartment in Cambridge. He envisioned the combination of information theory, neurophysiology, statistical mechanics, and computing machinery to understand the mystery of how the brain gives rise to the mind. Michael Arbib, who later became a research assistant in McCulloch's group, has recounted the influx of funding into this new area of research.

> *There was lots of money around so that being an RA was not particularly onerous. Basically, the Navy and other agencies gave lots of money to MIT and funneled them to various people and Warren [McCulloch] was one of the good guys so he had quite a lot of money to support bright young students.* (Anderson 216)

He further corroborated the opinions voiced by Lettwin regarding Pitts' intellectual prowess.

> *I think, where Pitts was the child and yet, in some ways, intellectually the more powerful of the pair though McCulloch knew an incredible amount about the brain and had been a very successful anatomist and still was at that time.* (Anderson 218)

During their time at MIT, the research trajectories of McCulloch and Pitts began to diverge. As recounted by Lettwin,

> *McCulloch became seduced into what can be done theoretically with nerve networks. Pitts by this time had more or less set himself against the concept of doing a synthetic job. To him it was much more important to come up with analytical notions of how such things were possible* (Anderson 9).

The McCulloch-Pitts model, although valuable in its own right, fell short in adequately capturing the intricacies of the biological brain, and thus it failed to elicit substantial enthusiasm among brain scientists. Recognizing this limitation, Wiener, drawing upon his mastery of statistics and probability theory, aimed to steer Pitts toward refining his brain model to attain a more realistic representation. Against this backdrop, Pitt's collaborations with Wiener gained substantial traction, and he started writing an extensive thesis delving into randomly connected probabilistic three-dimensional neural networks.

What transpired subsequently needs to be examined within the context of several influencing factors. As outlined in one of Norbert Wiener's biographies (Wiener 55), He was such a misfit in school that his father, Leo Wiener, a stringent Harvard languages professor, opted for homeschooling. Unfortunately, if Norbert fell short of expectations, he was occasionally labeled with derogatory terms such as "donkey," "brute," "ass," and "fool" in a multitude of languages—over forty, to be precise. These hurtful recollections remained a haunting presence throughout Norbert's lifetime. Despite these challenges, Norbert Wiener managed to complete his doctoral studies at Harvard by the age of eighteen, driven in the direction his father had encouraged. However, detaching himself from this imposed path took some time. Numerous sources shed light on Wiener's vulnerability to depression. An account from one of his MIT students notes that "his profound immersion in his own thoughts often rendered him unaware of his surroundings (Gangolli 772 )," Delving into Wiener's biography by Conway and Siegelm, a complex tapestry of attributes comes to the forefront (Conway). This includes a fusion of "astounding brilliance, a childlike sense of awe and trust, a humanist perspective stemming from resolute idealism, and, regrettably, his struggles with insecurity and a turbulent personal life."

Walter Pitts' father and brothers considered him an outsider. When he reached the age of 15, he took the significant step of running away from home, effectively severing all communication ties with his family (Smalheiser 217). It is noted that he held Warren McCulloch and Norbert Wiener in the light of paternal figures within his life (Anderson 9). Wilson, who extensively studied the complex dynamics between Pitts, McCulloch, Lettwin, and Wiener, arrives at the conclusion that

> *The affective inclinations in the group were perhaps too muddled and too muffled to withstand the force of conventional patriarchal fury, and Pitts was too fragile and too isolated to recover from the intellectual and emotional shock* (Wilson 847) resulting from the forthcoming incident.

Unexpectedly, Wiener severed all connections with the McCulloch group, including Pitts, sending shockwaves through their interactions. In a formal letter addressed to the President of MIT, he articulated a series of grievances against them, highlighting concerns such as the alleged misallocation of research funds. Speculation surrounding the true catalyst for this abrupt rupture has given rise to several theories (Anderson 9, Smalheiser 223). Nonetheless, in a biography penned by Conway and Siegelm, drawing from insights provided by Lettvin, an alternative narrative unfolds (Conway, 219). According to their account, Wiener's wife, Margaret, emerges as a pivotal figure in precipitating this division. Allegedly, she wove a fabricated tale out of her aversion to Wiener's association with the group, characterized by Bohemian inclinations and an unconventional lifestyle. Furthermore, it is posited that McCulloch's well-documented penchant for alcohol, and the extent to which he indulged in it, might have exacerbated Margaret's concerns due to Wiener's already weak emotional state. In essence, the reasons behind this sudden and drastic transformation in Wiener's relationships and alliances remain multi-faceted and open to interpretation.

This event left Pitts utterly devastated, bearing the brunt of its impact more than anyone else. Losing Wiener was akin to losing a father figure for him. Lettwin's account sheds light on the depth of this impact. Pitts had been preoccupied in the development of three-dimensional neural networks, a concept that was meticulously documented in his thesis, spanning hundreds of pages. Shockingly, he destroyed this painstaking work, a decision that Lettwin attributes to the emotional turmoil of the moment. The blow was so profound that, as Lettwin recalls, Pitts never managed to fully recover from it. Lettwin touchingly notes, "From that point on, there was no way of getting him interested in things (Anderson 9)." This continuing sadness endured until his tragic passing in 1969, seventeen years later. As recounted by Smalheiser, Pitts' response to the upheaval went beyond mere emotional distress (Smalheiser 223). Rather, he engaged in an unprecedented experiment with substances. Pitts, known for his exceptional intellect, took an unconventional path by synthesizing novel analogues of barbiturates and opiates within his laboratory. He delved further by subjecting himself to experiments involving long-chain alcohols, a testament to his complex coping mechanisms. Notably, in June 1954, Pitts' brilliance garnered recognition, as Fortune magazine distinguished him in its roster of Ten Top Young Scientists in U.S. universities (Smalheiser 223). Conway and Siegelm, in their analysis, underscore the pivotal role of the research trio—Wiener, Pitts, and McCulloch—and its subsequent dissolution. Their separation, according to the authors, emerges as a central reason for the unfulfilled potential of cybernetics. Conway and Siegelm lament that this division prevented cybernetics from achieving the remarkable success they believed it was destined for (Conway 233).

Cybernetics, as a discipline, was driven by a profound objective: to replicate the intricate workings of the human brain through computer hardware (Edwards 239). This ambition stemmed from their view of human intelligence as a dynamic interplay of information—an internal world of closed loops. In their eyes, intelligence emerged through the manipulation of information, a process empowered by the feedback loops. This perspective was rooted in the realization that intelligence, whether displayed by humans or other systems, wove information processing and adaptive responses in the pursuit of objectives. Their approach revolved around the creation of self-organizing machines poised to attain complex behaviors by engaging with their surroundings—a representation of closed-loop dynamics.

**The Epoch of Symbolic AI: Pioneering Intelligence Using Software**

The subsequent wave of progress emerged through a group of researchers whose perception of computers went beyond their pragmatic utility. They regarded these machines not merely as instruments for pragmatic problem-solving, but rather as automated representations of mathematical models with profound intellectual attraction. This intellectual effort resulted in the form of Artificial Intelligence (AI), a term formally coined in 1956 (Crevier 50). Departing from the ambition to simulate cognitive functions through hardware replication, AI pursued a different trajectory by attempting to exhibit intelligent behavior within software constructs. Scientists such as Allen Newell, Herbert A. Simon, Marvin Minsky, and John McCarthy stand among the pioneers of AI research. Their contributions paved the way for the first wave of artificial intelligence (Crevier 32-44).

The 1956 Dartmouth Conference, a summer school held at Dartmouth College in Hanover, New Hampshire, is widely acknowledged as the pivotal origin of AI as an academic discipline (Crevier 48). This two-month event, co-convened by Marvin Minsky and John McCarthy, aimed to explore the notion that all aspects of learning and intelligence could be comprehensively explained to the extent that machines could replicate them precisely. The documented discussions reveal that Minsky emphasized topics like learning theory and the necessity for precise descriptions of the principles behind the brain's physiological structure as significant focal points during the gathering (Penn 172).

One of McCarthy's objectives was to create a hybrid logical and natural language for AI, aiming to provide machines with a foundational comprehension of the world. Between 1956 and 1958, he successfully realized this goal through the development of a programming language named LISP (Penn 154). LISP manipulates lists and programs written in LISP are inherently structured as lists themselves. This language, which emerged as the lingua franca of symbolic AI in its early stages, enabled the field to make significant strides.

The advent of digital computers in the 1950s and the subsequent widespread adoption of high-level programming languages played a pivotal role in advancing and shaping symbolic AI. These programming languages introduced a higher level of abstraction, enabling researchers to focus on directly translating symbolic and logical concepts into code. Symbolic AI revolves around the manipulation of explicit symbols and rules to represent knowledge and perform reasoning tasks. In symbolic AI, knowledge is typically conveyed through symbols, logical statements, and rule-based systems. The primary objective is to manipulate these symbols to deduce conclusions and solve problems. Symbolic AI systems are rule-driven, relying on formal logic for reasoning. They excel in well-structured and clearly defined domains, where explicit rules and logical relationships can be easily articulated. One of the early achievements of symbolic AI was the creation of expert systems, which encoded human expertise and knowledge in the form of rules to solve specific problems. However, symbolic AI has inherent limitations when it comes to handling uncertainty and processing vast amounts of unstructured data, which makes it less suitable for tasks like image recognition and natural language understanding.

As later elaborated in this essay, the early AI pioneers aimed to establish a distinct identity separate from cyberneticians. However, it is important to note that this perspective is not entirely accurate. What is intriguing is that the impact of cybernetics on this group of researchers remained largely unexplored until the publication of Paul Edwards' book, "The Closed World," in 1996 (Edwards 239).

Among the influential AI researchers of the time were Allen Newell and Herbert A. Simon, creators of the Logic Theorist in 1956, a program capable of proving mathematical theorems using symbolic logic. Notably, Oliver Selfridge, a prominent disciple of Norbert Wiener and recognized as the "Father of Machine Perception," exerted a significant influence on Newell's intellectual journey. Selfridge's pioneering programs occupied the intersection of cybernetics and symbolic information processing, representing a pivotal transitional phase in the evolution of computational models. This transitional quality of Selfridge's work not only resonated with Newell's own intellectual inclinations but also provided him with a conceptual framework to bridge the gap between biologically inspired ideas and symbolic AI approaches (Edwards 250).

Marvin Minsky was one of the early proponents of symbolic AI and the development of expert systems. He believed that intelligence could be replicated through the manipulation of symbols and logical rules. His contributions were integral in shaping the nascent stages of the field's evolution. Nonetheless, Jonathan Penn has highlighted a noteworthy aspect regarding Minsky's academic journey. During his tenure as a doctoral researcher in mathematics at Princeton University in 1950, Minsky extensively immersed himself in cybernetic theory (Penn 160). It is worth noting that Minsky's transition toward symbolic reasoning commenced around 1954, a mere two years prior to the pivotal Dartmouth Conference that sought to establish the roadmap for AI research.

**The Dawn of Neural Networks: Revolutionizing Intelligence**

Frank Rosenblatt, an AI researcher deeply engaged in learning theory at Cornell University during that period, was conspicuously absent from the list of invitees to the Dartmouth Conference (Penn 140). In 1957, Rosenblatt made a groundbreaking contribution by publishing a technical report in 1957 and a pivotal paper a year later on "Perceptrons," a term he introduced which now corresponds to what we commonly refer to as neural networks (Rosenblatt). Combining his background in psychology with a strong reliance on statistical methodologies, Rosenblatt's perceptron project was centered around the concept of training a machine through associative logic. His work marked the inception of the first model capable of acquiring weights from examples, thereby advancing the concept of learning through practice. The initial form of the model emerged as a simulation within an IBM 704 computer (Penn 82). Rosenblatt's model, while admittedly a simplification of the nervous system, was rooted in the foundational framework of McCulloch and Pitts neurons. This approach, inspired by the intricacies of the brain, signified a marked deviation from the trajectory pursued by the symbolic AI community of that era. The fundamental distinction lay in how symbolic AI perceived knowledge: as a hierarchical system of predefined rules and procedures. In contrast, the approach taken by perceptron research embraced a perspective where knowledge was acquired organically, emerging from intricate interactions with the environment, and developing from the ground up (Penn 82).

Rosenblatt emerged as not only a brilliant scientist but also a captivating figure with a knack for media navigation. As mentioned by Mikel Olazaran, he possessed qualities that would make a press agent's dreams come true (Olazaran 105). Interestingly, in contrast to the unfolding events that lay ahead, his earlier years held an intriguing connection: during their time as fellow students at the Bronx High School of Science, he shared a friendship with Marvin Minsky that traced back to their childhood days (Gravier 102).

The scientific community's frustration originated from Rosenblatt's presentation, marked by a unique flourish, of his work to the media. This was followed by the unfortunate misrepresentation of his findings in the subsequent reporting. One such example was how the prestigious Science magazine featured a headline titled "Human Brains Replaced?" that suggested "Perceptron may eventually be able to learn, make decisions, and translate languages." (Gravier 103) The New York Times covered an event involving a primary sponsor of the project

*The Navy revealed the embryo of an electronic computer today that it expects will be able to walk, talk, see, write, reproduce itself and be conscious of its existence. Later perceptrons will be able to recognize people and call out their names and instantly translate speech in one language to speech and writing in another language* (Olazaran 100).

The New Yorker also quoted Rosenblatt as expressing:

*The Perceptron can tell the difference between a dog and a cat, though so far, according to our calculations, it would not be able to tell whether the dog was to the left or to the right of the cat. We still have to teach it depth perception and refinements of judgment* (Gravier 103).

The challenge of image classification under various demanding conditions persisted as a formidable problem for decades to follow. It was only within the past decade that neural networks successfully reached these significant milestones. Critics leveled accusations at Rosenblatt, asserting that he had not upheld scientific standards and had instead employed the media in a biased manner (Olazaran 103). Nevertheless, when addressing scientific audiences, Rosenblatt exhibited caution in linking his work to prior research.

Minsky, a prominent advocate of the symbolic AI approach during that era, had actually delved into his own experimentation with neural networks while at Harvard. Additionally, he engaged in thorough theoretical analysis during his tenure at Princeton. While Rosenblatt's work stands as the most renowned non-symbolic AI project of its time, it was part of a broader trend that commenced in the early 1950s and gained momentum leading up to the late 1950s. Minsky was among some of the prominent scientists who were concerned about this growing trend. Later he claimed that

> *schemes quickly took root, and soon there were perhaps as many as a hundred groups, large and small, experimenting the model either as a 'learning machine' or in the guise of 'adaptive' or 'self-organizing' networks or 'automatic control' systems*. (Olazaran 110).

Advocates of the neural network approach readily acknowledged its limitations. Foremost among these constraints was its single-layer architecture, which rendered it incapable of executing numerous essential functions. During that period, the absence of a methodology to train multilayer networks hindered the practical utility of this algorithm. The proponents of this approach asserted that single-layer networks represented just the initial stage, and while acknowledging their significant limitations, they remained confident that more intricate systems would eventually surmount these challenges. However, these seemingly exaggerated claims about its potential and capabilities sparked numerous, often intense debates within scientific circles. Among those voicing skepticism was Minsky, who prominently led discussions against the perceptron approach. Amidst the fervent debates of the 1960s, a particularly heated period, Marvin Minsky and Seymour Papert, both associated with MIT at the time, undertook a noteworthy and resource-intensive endeavor. They chose to 're-enact' the perceptron's outcomes, a meticulous process involving replicating every step the original author had taken. Unavoidably, the undertaking turned out to be a protracted endeavor, surpassing the initially anticipated timeframe. The culmination of this effort, along with its subsequent analysis, eventually saw the light of day in 1969 with the release of an unpublished technical manuscript and the publication of a revised and de-venomized book titled "Perceptrons (Minsky)."

It is widely acknowledged that the critique presented in the book, coupled with the influential stature of Minsky and Papert during that era, exerted significant influence in temporarily stalling the progress of neural-network research in the United States (Olazaran 183). This pause in advancement saw Symbolic AI reclaim its previously dominant position, maintaining its supremacy until the resurgence of neural network research in the 1980s.

In hindsight, Papert later conceded that this redirection was largely a mistake, given that nearly half of the findings presented in the book actually supported the potential of Perception. Similarly, Minsky later acknowledged that the book might have been an "overkill." (Bernstein) It is widely acknowledged that the initial optimism surrounding the advancement of AI may have been excessive. Nevertheless, it is equally important to recognize that the backlash against neural networks during this time may have been too extreme.

Today, it is worth noting that neural networks have become the driving force behind the vast majority of AI success stories, underscoring the remarkable turnaround in perception and the undeniable impact they have had on the field of artificial intelligence.

**Unifying the Pinnacle of Intelligence: Neuro-Symbolic AI**

A fusion of philosophical, historical, and social influences has often led to the conventional belief that symbolic AI and neural network approaches were fundamentally and irrevocably separate entities (Schneider) . However, Vasant Honavar has compellingly challenged this seemingly insurmountable divide (Honavar). He has emphasized the alignment between the foundational philosophical assumptions and scientific hypotheses that have molded both approaches in the realm of modeling cognition and engineering intelligent systems. He has pointed out that both approaches share the core working hypothesis that cognition, or the processes of thought, can, at some level, be effectively modeled through computation.

Neural networks have indeed demonstrated remarkable proficiency in processing and discerning patterns from raw data. Nevertheless, they often lack the explicit representations of background knowledge essential for tasks such as abstraction, analogical reasoning, and long-term planning. In contrast, symbolic knowledge-based AI excels in modeling knowledge, facilitating traceability, and enabling auditability of AI system decisions. The emerging neuro-symbolic paradigm seeks to harmonize and synthesize the strengths of both approaches, presenting a highly promising avenue for advancing artificial intelligence by enhancing its capacity for explainability and causal reasoning (Sheth).

A pressing question that naturally arises is why this integration was not ventured into at an earlier juncture. I contend that one of the primary impediments to the exploration of integrating these two approaches lies in the enduring historical schism that persisted between these two scientific communities. This divide made collaborative endeavors challenging during the formative stages of development, ultimately resulting in the establishment of two parallel streams of research.

**Conclusions**

It is natural for pioneers in a field to be concerned with documenting their own history. Typically, such efforts focus on developing a historical narrative from an intellectual standpoint. However, as the SCOT framework suggests, the emergence and evolution of new fields are the outcomes of a complex interplay among technology, societal factors, and human agency.

Some of the foundational ideas of cybernetics found their roots in the Macy Conferences, a series of multidisciplinary gatherings held between 1946 and 1953, supported by the Macy Foundation.

In the wake of World War II, numerous scientists had returned from participation in multidisciplinary military research projects, and their achievements were held in high regard. For instance, Norbert Wiener had incorporated feedback control systems into anti-aircraft gun fire control systems to enhance targeting accuracy during the war. Consequently, this generation of scientists possessed valuable experience in engaging in multidisciplinary research projects, making it relatively easier to replicate such collaborative efforts in academic settings.

This contrasts with the relatively unproductive Dartmouth Conference (Minsky), which coined the term "Artificial Intelligence" and took place a decade later in a different socio-political context. Moreover, the cybernetics movement gained momentum at a time when some of the scientific and technical advances of the war years—such as the modern general-purpose computer and models based on it—were just becoming publicly available. This occurred within the broader context of postwar practical requirements, political discussions, and social networks, all of which played a pivotal role in shaping the trajectory of cybernetics.

A significant portion of pioneering cybernetics research was carried out by a close-knit group of scientists who worked together within the same university for an extended period. This group, which included luminaries like McCulloch, Wiener, Pitts, and Lettvin, not only collaborated professionally but also shared strong social bonds. McCulloch, in particular, stood out as an intellectually open, charismatic, warm, and personally informal figure. His hospitality and generosity extended to many young scientists, including one noteworthy example, Pitts. He, a prodigious talent, was homeless, in need, shy, and somewhat eccentric. The remarkably productive collaboration between McCulloch and Pitts might never have transpired without the emotional support McCulloch provided to Pitts.

On the other hand, Wiener, another influential 'father' figure in Pitts's professional and personal life, presented a contrasting personality. Wiener, though brilliant, was socially insecure and awkward. Unfortunately, their close social ties eventually contributed to the breakdown of their scientific collaborations, resulting in a tragic personal loss and a setback to the progress of science. It's worth noting that Pitts, initially an outsider, had been greatly aided by the support of these insiders, ultimately benefiting the field of science.

Rosenblatt, in the years that followed, would build upon the McCulloch and Pitts model to develop his ideas of perceptrons, which marked the foundational point for the neural networks we use today. However, by the time of the relationship breakdown, Pitts was already working on an improved version of the model. Lettvin, who enjoyed a close professional and personal association with Walter Pitts during that era, provided insightful commentary on Pitts' pioneering work. He remarked, "Walter was ahead of his time. He envisioned a layered device, a three dimensional net..nobody else was tackling it.. and got some very interesting results (Conway 232-233)." One can only imagine the potential implications if this enhanced model had been available to Rosenblatt at that crucial juncture.

What's evident is that exceptional individuals are necessary to pioneer new scientific fields. Beyond intelligence and creativity, they must also possess the skills to secure funding, gain media attention, and build teams. Some of these unconventional individuals may be highly sensitive, and this essay illustrates how social factors can hinder their progress.

The majority of modern general-purpose computers are often referred to as Von Neumann machines, owing to their foundational reliance on the stored-program concept proposed by John von Neumann, a prominent figure in the field of cybernetics. This architectural paradigm, outlined in his seminal work "First Draft of a Report on the EDVAC" in 1945, cited only one published report, the 1943 McCulloch-Pitts paper (Conway 150). However, Conway and Siegelman have postulated that John von Neumann may have been exposed to certain visionary ideas through his close collaboration with Norbert Wiener. He had submitted forward-thinking suggestions to Vannevar Bush, the presidential science advisor at the time, in 1940, which proposed five key features in the EDVAC's eventual architecture. They even cite informed sources of the era who assert that "Most of the elements of the Von Neumann machine, save the stored program, are present in Wiener's memorandum," and suggest that, had Bush circulated Wiener's memorandum widely, we might now refer to the "Wiener-Von Neumann" or even the "Wiener machine" instead of the "Von Neumann machine (Conway 151)."

This historical narrative, intertwined with the inception of computer architecture, underscores the interpretive flexibility that has often been exploited in attributing credit, particularly in the nascent stages of fields like Artificial Intelligence. The early development of AI spanned diverse academic departments, and its findings were disseminated through various publication channels, given that it had not yet crystallized into a well-defined discipline.

One illustrative case is that the debate revolves around the attribution of credit for the development of the backpropagation algorithm, which is now a cornerstone of contemporary deep learning. In this complex historical narrative, several individuals have been associated with its invention, each with their own claims. One perspective attributes the pioneering work to Paul Werbos, who, as a mathematics PhD student at Harvard in the early 1970s, devised a technique known as dynamic-feedback for neural network-type models (Werbos). Another viewpoint credits Seppo Linnainmaa, who introduced a similar concept in 1970 but without direct reference to neural networks (Schmidhuber). Additionally, David Rumelhart, who independently reinvented the algorithm in 1986, claimed ignorance of Werbos' contributions and coined the term "backpropagation algorithm," thus adding another layer to the controversy (Synced). This complex debate serves as a testament to the intricate nature of emerging fields. It is crucial to recognize the historical context within which Werbos and Linnainmaa conducted their research. At the time, neural networks as a discipline were in their infancy, lacking the development and prominence they enjoy today. Consequently, these early innovators did not explicitly position their work within the neural network domain, partly due to limited interest and practical challenges. Moreover, the technological landscape of the era posed significant hurdles.

Computers were markedly slower than today's counterparts, impeding the efficient implementation and validation of their algorithms. For instance, Paul Werbos encountered difficulties in convincing his thesis committee at Harvard, as skepticism about the algorithm's validity prevailed. He was even advised to seek the opinion of someone with more expertise and credibility. In a telling example, Werbos turned to Marvin Minsky for counsel, only to receive a lukewarm response (Olazaran 248). Nearly two decades later, Minsky retrospectively justified his skepticism by citing the slow convergence of the algorithm (Olazaran 249), which was not surprising given the state of computing technology at the time. This example underscores issues of historical recognition and the challenges faced by early pioneers in an emerging field, where limited resources and understanding often hindered the full appreciation of groundbreaking contributions.

As cited by John von Neumann, his architectural framework drew inspiration from the McCulloch-Pitts model (Conway 150). However, this model harbored certain imperfections. It is worth noting that had these flaws been rectified, a scenario that Walter Pitts was actively pursuing before an unfortunate decision to destroy his own PhD thesis, both the Von Neumann architecture underpinning contemporary computers and the present-day neural networks, which were inspired by the McCulloch-Pitts model, might have attained even greater levels of sophistication and effectiveness.

The clash between symbolic AI and neural network approaches unfolded within a distinct socio-technological landscape compared to the era of cybernetics. During this period, key stakeholders, including scientists, funding agencies, the media, and the broader public, had become increasingly aware of the emerging AI innovations. By the time perceptrons and neural networks began to gain prominence, the leaders of the symbolic AI movement had already solidified their pioneer status and earned considerable prestige. They occupied the position of insiders, whereas the neural network researchers found themselves on the outside. Their reactions were motivated by several factors, including the aspiration to steer the narrative and shape the trajectory of technology, fierce competition for funding, a determination to challenge what they perceived as unverified assertions, and a degree of exasperation with the manner in which these claims were being presented.

Creating a fresh narrative marks a pivotal milestone in pioneering a nascent domain. As exemplified by McCarthy, one of the AI pioneers, the inception of the term 'artificial intelligence' was driven by the desire to disentangle from the web of 'cybernetics.' He notably stated, "I wished to avoid having either to accept Wiener as a guru or having to argue with him (Penn 131)." However, once such distinction is attained, safeguarding it takes precedence. As pioneers, Minsky and Papert were eager to encourage more researchers to align with their approach and grew concerned as neural network research gained traction. They sought to halt what they perceived as an unwarranted diversion of resources into an area they deemed scientifically and practically questionable. The rhetoric was to "kill the perceptron (Olazaran 168)." Their objective was to restore the equilibrium of AI funding and research in favor of their own approach.

The current AI hype, to some extent, is substantiated by tangible achievements. Yet, during the 1960s, the discourse predominantly revolved around the prospects and potential of various approaches. This created more room for interpretive flexibility to play a significant role in the debates. This meant that the way technology and its potential were presented to funders, the media, and the public carried greater weight. Harnessing the power of interpretive flexibility required crafting the technology narrative in a manner that resonated with the values held by the target audience. Despite the fact that the specific criticism primarily targeted single-layer perceptrons on an objective basis, it had a broader, negative undertone directed towards the entire neural network paradigm. Owing to the sway of these influential voices, it was this overall critical tone that gained prominence and influence.

This contributed to an environment teeming with heightened emotions and tensions, leaving behind a sustained, negative impact. The symbolic AI community and the neural networks community were often seen as two distinct camps. This schism had a significant impact, with relentless criticism directed towards the neural network approach casting a pall over research endeavors, commercial enthusiasm, and funding prospects. Rosenblatt, a prominent figure in the neural network domain, relied heavily on financial support from the Office of Naval Research (ONR) and Advanced Research Projects Agency (ARPA). Regrettably, the pervasive skepticism surrounding the potential of neural networks played a pivotal role in dwindling funding opportunities for many neural network projects. In the context of unrelenting criticism, declining funding, and the notable shift of key commercial supporters to alternative approaches, coupled with influential researchers transitioning to different domains, we can view this complex landscape through the lens of Actor Network Theory (ANT). These interconnected events created a network effect, culminating in a situation where, when the decisive attack in the form of the publication of the perceptrons book occurred, Rosenblatt found himself inadequately positioned to mount a robust defense, primarily due to a lack of supportive alliances.

From a retrospective, objective perspective, it raises questions as to why previous efforts were not undertaken to bridge the gap between symbolic AI and neural networks. It underscores that technological progress is not solely a product of technical feasibility; rather, it is molded by the intricate interplay of social dynamics, technological advancements, institutional influences, and human choices. History vividly illustrates how the network effects generated by these diverse actors often lead to winners and losers. During the symbolic AI's ascendancy, the neural network movement found itself on the losing side, and vice versa. This polarity did little to foster a collaborative environment conducive to exploring groundbreaking unified architectures. Each camp stuck to its chosen path, even though both faced unique challenges.

However, in an era unburdened by the baggage of past debates, it appears that the time has arrived for unrestricted exploration of unified approaches.